\documentclass[11pt,a4paper]{article}
\usepackage[hyperref]{acl2020}
\usepackage{microtype}
\usepackage{subfigure}
\usepackage{textcomp}
\usepackage{latexsym}
\usepackage{graphicx}
\usepackage{enumitem}
\usepackage{multirow}
\usepackage{amssymb}
\usepackage[normalem]{ulem}
\usepackage{times}
\usepackage{ulem}
\usepackage{url}
\usepackage[hang,flushmargin]{footmisc}  
\setenumerate[1]{leftmargin=*}         
\setitemize[1]{leftmargin=*}           

\aclfinalcopy 

\newcommand{\textsec}[1]{\textsection\ref{#1}}
\newcommand{\LN}{\linebreak\noindent}

\title{Transformers to Learn Hierarchical Contexts in Multiparty Dialogue\\for Span-based Question Answering}

\author{Changmao Li \\
  Department of Computer Science \\
  Emory University \\
  Atlanta, GA, USA \\
  \texttt{changmao.li@emory.edu} \\\And
  Jinho D. Choi\\
  Department of Computer Science \\
  Emory University \\
  Atlanta, GA, USA \\
  \texttt{jinho.choi@emory.edu} \\}
\date{}
\begin{document}
\maketitle

\begin{abstract}
We introduce a novel approach to transformers that learns hierarchical representations in multiparty dialogue.
First, three language modeling tasks are used to pre-train the transformers, token- and utterance-level language modeling and utterance order prediction, that learn both token and utterance embeddings for better understanding in dialogue contexts.
Then, multi-task learning between the utterance prediction and the token span prediction is applied to fine-tune for span-based question answering (QA).
Our approach is evaluated on the \textsc{FriendsQA} dataset and shows improvements of 3.8\% and 1.4\% over the two state-of-the-art transformer models, \texttt{BERT} and \texttt{RoBERTa}, respectively.

\end{abstract}
\section{Introduction}
\label{sec:introduction}

Transformer-based contextualized embedding approaches such as \texttt{BERT} \cite{devlin_2019}, \texttt{XLM} \cite{lample_2019}, \texttt{XLNet} \cite{yang_2019a}, \texttt{RoBERTa} \cite{liu_2019}, and \texttt{AlBERT} \cite{lan_2019} have re-established the state-of-the-art for practically all question answering (QA) tasks on not only general domain datasets such as \textsc{SQuAD} \cite{Rajpurkar_2016,Rajpurkar_2018}, \textsc{MS Marco} \cite{bajaj_2016}, \textsc{TriviaQA} \cite{Joshi_2017}, \textsc{NewsQA} \cite{Trischler_2017}, or \textsc{NarrativeQA} \cite{Ko_isk__2018}, but also multi-turn question datasets such as \textsc{SQA} \cite{iyyer_2017}, \textsc{QuAC} \cite{Choi_2018}, \textsc{CoQA} \cite{Reddy_2019}, or CQA \cite{Talmor_2018}.
However, for span-based QA where the evidence documents are in the form of multiparty dialogue, the performance is still poor even with the latest transformer models \cite{Sun_2019,yang_2019}  due to the challenges in representing utterances composed by heterogeneous speakers.

Several limitations can be expected for language models trained on general domains to process dialogue.
First, most of these models are pre-trained on formal writing, which is notably different from colloquial writing in dialogue; thus, fine-tuning for the end tasks is often not sufficient enough to build robust dialogue models.
Second, unlike sentences in a wiki or news article written by one author with a coherent topic, utterances in a dialogue are from multiple speakers who may talk about different topics in distinct manners such that they should not be represented by simply concatenating, but rather as sub-documents interconnected to one another.


This paper presents a novel approach to the latest transformers that learns hierarchical embeddings for tokens and utterances for a better understanding in dialogue contexts.
While fine-tuning for span-based QA, every utterance as well as the question are separated encoded and multi-head attentions and additional transformers are built on the token and utterance embeddings respectively to provide a more comprehensive view of the dialogue to the QA model.
As a result, our model achieves a new state-of-the-art result on a span-based QA task where the evidence documents are multiparty dialogue.
The contributions of this paper are:\footnote{All our resources including the source codes and the dataset with the experiment split are available at\\\url{https://github.com/emorynlp/friendsqa}}


\begin{itemize}[itemsep=0.1em]
\item New pre-training tasks are introduced to improve the quality of both token-level and utterance-level embeddings generated by the transformers, that better suit to handle dialogue contexts (\textsec{ssec:pretraining}). 
\item A new multi-task learning approach is proposed to fine-tune the language model for span-based QA that takes full advantage of the hierarchical embeddings created from the pre-training (\textsec{ssec:finetuning}).
\item Our approach significantly outperforms the previous state-of-the-art models using \texttt{BERT} and \texttt{RoBERTa} on a span-based QA task using dialogues as evidence documents  (\textsec{sec:experiments}).

\end{itemize}

\begin{figure*}[htbp!]
\centering

\subfigure[Token-level MLM (\textsec{sssec:pretraining-1})]
{\label{fig:pretraining-1}\includegraphics[scale=0.35]{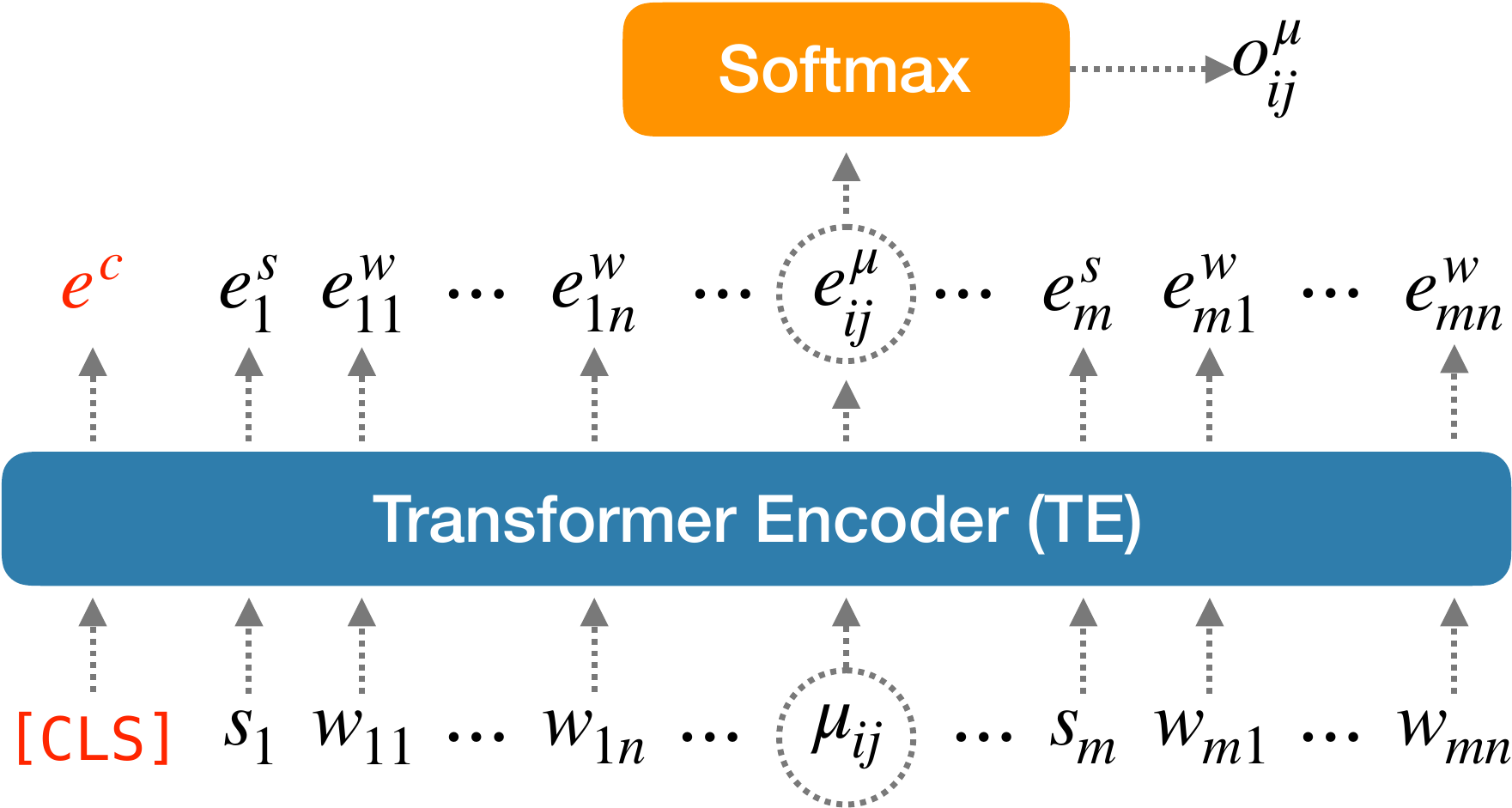}}
~~~~~~~~~~~~~~~~~
\subfigure[Utterance-level MLM (\textsec{sssec:pretraining-2})]
{\label{fig:pretraining-2}\includegraphics[scale=0.35]{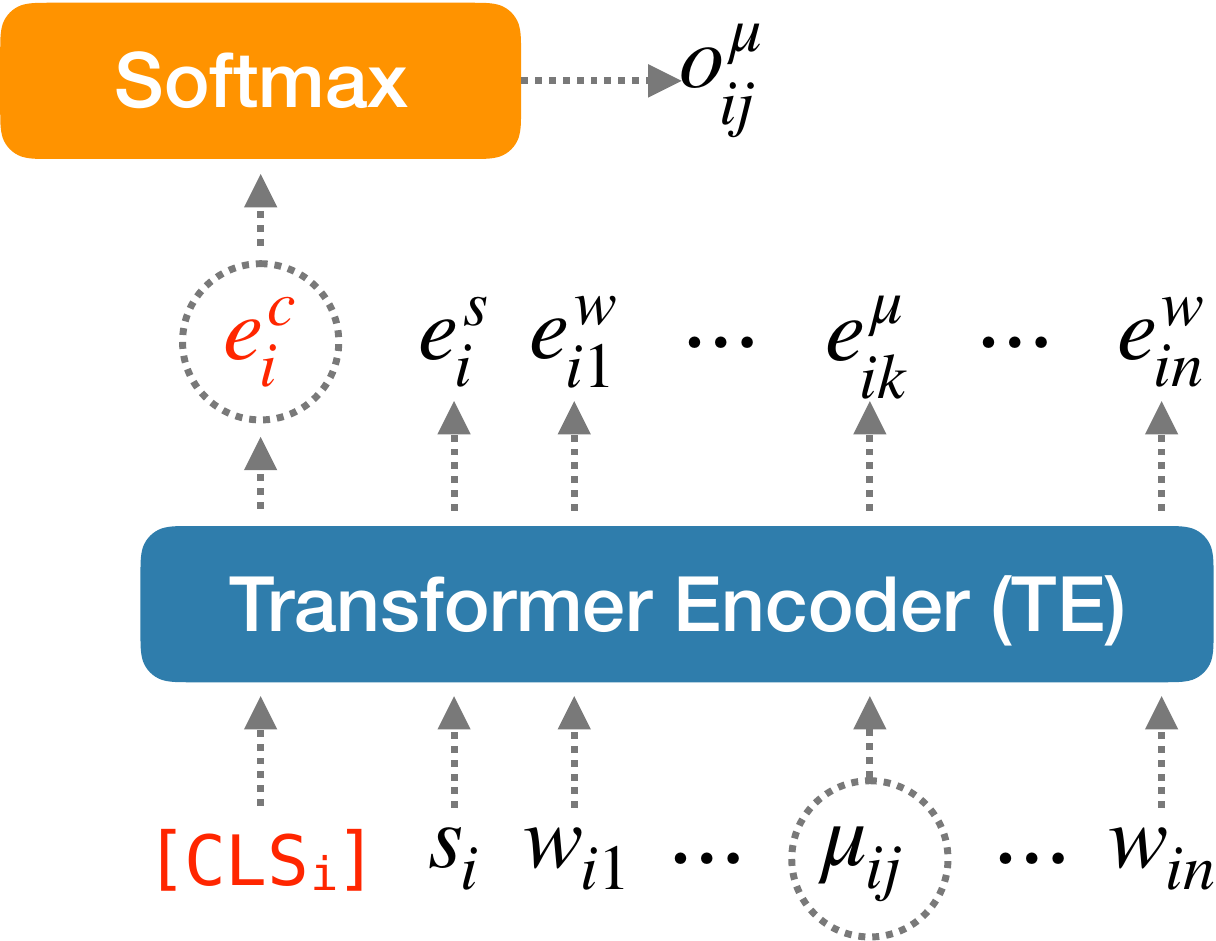}}
\\
\subfigure[Utterance order prediction (\textsec{sssec:pretraining-3})]
{\label{fig:pretraining-3}\includegraphics[scale=0.35]{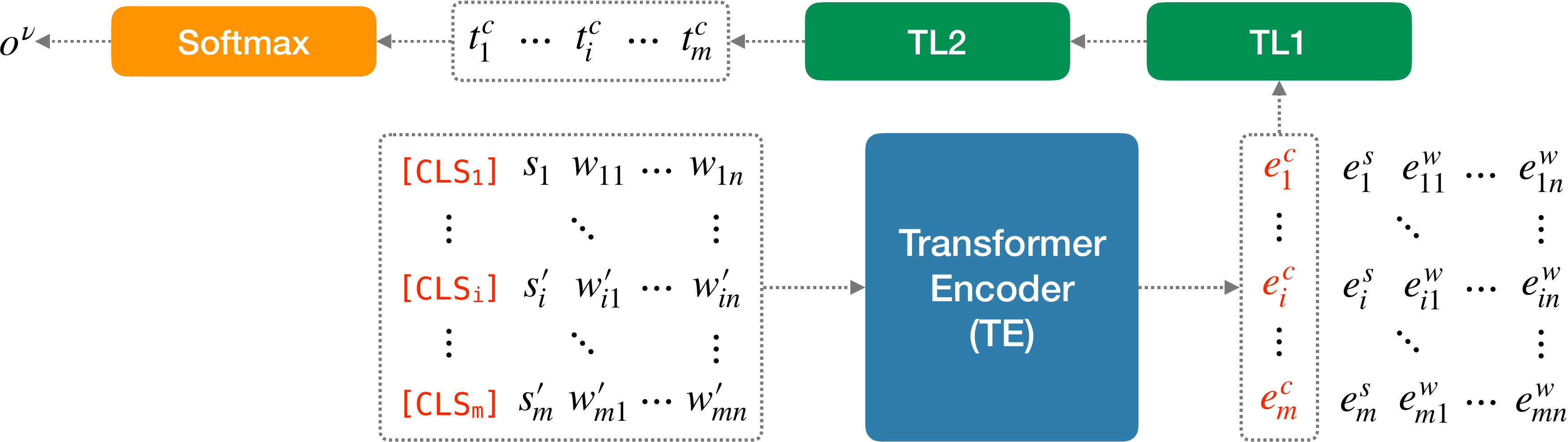}}

\caption{The overview of our models for the three pre-training tasks (Section~\ref{ssec:pretraining}).}
\label{fig:pretraining}
\end{figure*}


\section{Transformers for Learning Dialogue}
\label{sec:approach}

This section introduces a novel approach for pre-training (Section~\ref{ssec:pretraining}) and fine-tuning (Section~\ref{ssec:finetuning}) transformers to effectively learn dialogue contexts.
Our approach has been evaluated with two kinds of transformers, BERT \cite{devlin_2019} and RoBERTa \cite{liu_2019}, and shown significant improvement to a question answering task (QA) on multiparty dialogue (Section~\ref{sec:experiments}).


\subsection{Pre-training Language Models}
\label{ssec:pretraining}

Pre-training involves 3 tasks in sequence, the token-level masked language modeling (MLM; \textsec{sssec:pretraining-1}), the utterance-level MLM (\textsec{sssec:pretraining-2}), and the utterance order prediction (\textsec{sssec:pretraining-3}), where the trained weights from each task are transferred to the next task.
Note that the weights of publicly available transformer encoders are adapted to train the token-level MLM, which allows our QA model to handle languages in both dialogues, used as evidence documents, and questions written in formal writing. 
Transformers from BERT and RoBERTa are trained with static and dynamic MLM respectively, as described by \citet{devlin_2019,liu_2019}.


\subsubsection{Token-level Masked LM}
\label{sssec:pretraining-1}

Figure~\ref{fig:pretraining-1} illustrates the token-level MLM model.
Let $D = \{U_1, \ldots, U_m\}$ be a dialogue where $U_i = \{s_i, w_{i1}, \ldots, w_{in}\}$ is the $i$'th utterance in $D$, $s_i$ is the speaker of $U_i$, and $w_{ij}$ is the $j$'th token in $U_i$.
All speakers and tokens in $D$ are appended in order with the special token \texttt{CLS}, representing the entire dialogue, which creates the input string sequence $I = \{\texttt{CLS}\} \oplus U_1 \oplus \ldots \oplus U_n$.
For every $w_{ij} \in I$, let $I^\mu_{ij} = (I \setminus \{w_{ij}\}) \cup \{\mu_{ij}\}$, where $\mu_{ij}$ is the masked token substituted in place of $w_{ij}$.
$I^\mu_{ij}$ is then fed into the transformer encoder (\texttt{TE}), which generates a sequence of embeddings $\{e^c\} \oplus E_1 \oplus \ldots \oplus E_m$ where $E_i = \{e^s_i, e^w_{i1}, .., e^w_{in}\}$ is the embedding list for $U_i$, and $(e^c, e^s_i, e^w_{ij}, e^\mu_{ij})$ are the embeddings of $(\texttt{CLS}, s_i, w_{ij}, \mu_{ij})$ respectively.
Finally, $e^\mu_{ij}$ is fed into a softmax layer that generates the output vector $o^\mu_{ij} \in \mathbb{R}^{|V|}$ to predict $\mu_{ij}$, where $V$ is the set of all vocabularies in the dataset.\footnote{$n$: the maximum number of words in every utterance,\\$m$: the maximum number of utterances in every dialogue.}



\begin{figure*}[htbp!]
\centering
\includegraphics[scale=0.35]{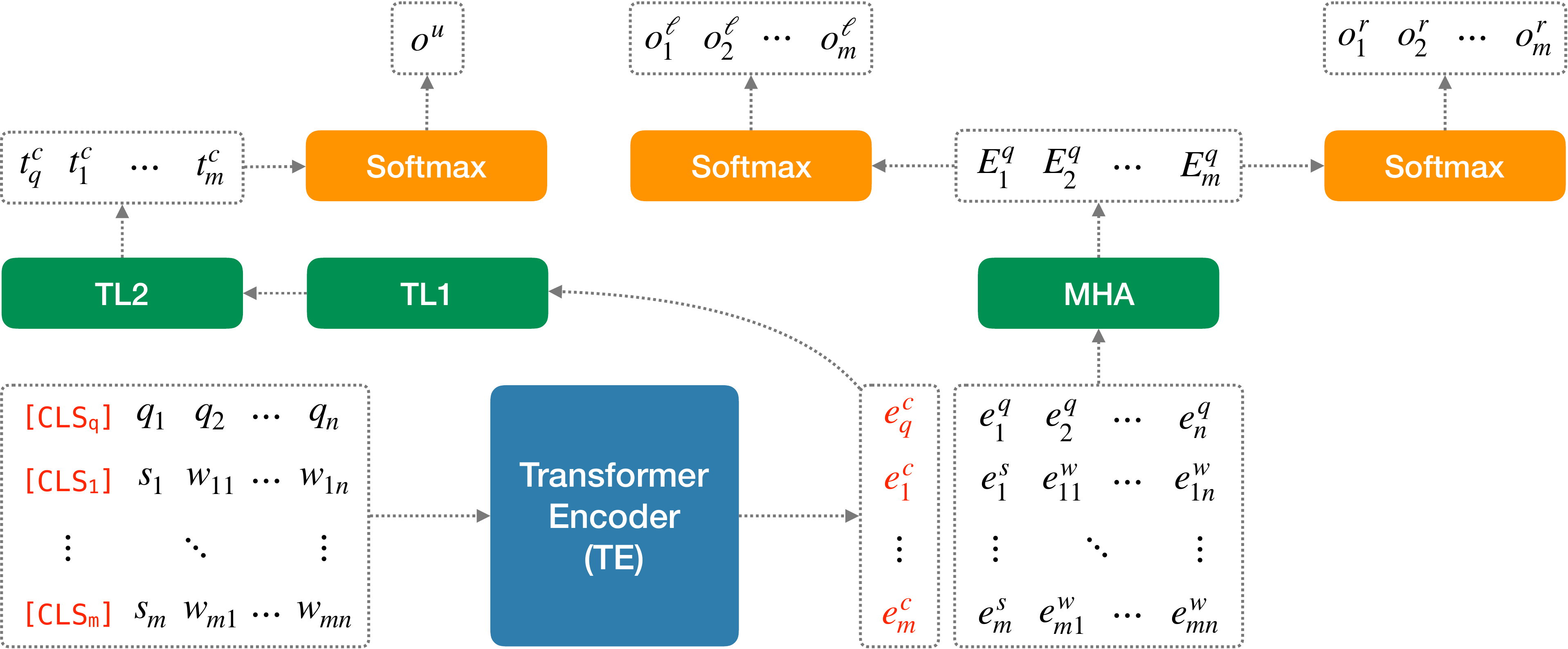}
\caption{The overview of our fine-tuning model exploiting multi-task learning (Section~\ref{ssec:finetuning}).}
\label{fig:finetuing}
\end{figure*}

\subsubsection{Utterance-level Masked LM}
\label{sssec:pretraining-2}

The token-level MLM (t-MLM) learns attentions among all tokens in $D$ regardless of the utterance boundaries, allowing the model to compare every token to a broad context; however, it fails to catch unique aspects about individual utterances that can be important in dialogue.
To learn an embedding for each utterance, the utterance-level MLM model is trained (Figure~\ref{fig:pretraining-2}).
Utterance embeddings can be used independently and/or in sequence to match contexts in the question and the dialogue beyond the token-level, showing an advantage in finding utterances with the correct answer spans (\textsec{sssec:finetuning-1}).

\noindent For every utterance $U_i$, the masked input sequence $I^\mu_{ij} = \{\texttt{CLS}_i\} \oplus \{(U_i \setminus \{w_{ij}\}) \cup \mu_{ij}\}$ is generated.
Note that \texttt{CLS}$_i$ now represents $U_i$ instead of $D$ and $I^\mu_{ij}$ is much shorter than the one used for t-MLM.
$I^\mu_{ij}$ is fed into \texttt{TE}, already trained by t-MLM, and the embedding sequence $E_i = \{e^c_i, e^s_i, e^w_{i1}, .., e^w_{in}\}$ is generated.
Finally, $e^c_i$, instead of $e^\mu_{ij}$, is fed into a softmax layer that generates $o^\mu_{ij}$ to predict $\mu_{ij}$.
The intuition behind the utterance-level MLM is that once $e^c_i$ learns enough contents to accurately predict any token in $U_i$, it consists of most essential features about the utterance; thus, $e^c_i$ can be used as the embedding of $U_i$.



\subsubsection{Utterance Order Prediction}
\label{sssec:pretraining-3}

The embedding $e^c_i$ from the utterance-level MLM (u-MLM) learns contents within $U_i$, but not across other utterances.
In dialogue, it is often the case that a context is completed by multiple utterances; thus, learning attentions among the utterances is necessary.
To create embeddings that contain cross-utterance features, the utterance order prediction model is trained (Figure~\ref{fig:pretraining-3}).
Let $D = D_1 \oplus D_2$ where $D_1$ and $D_2$ comprise the first and the second halves of the utterances in $D$, respectively.
Also, let $D' = D_1 \oplus D'_2$ where $D'_2$ contains the same set of utterances as $D_2$ although the ordering may be different.
The task is whether or not $D'$ preserves the same order of utterances as $D$.

For each $U_i \in D'$, the input $I_i = \{\texttt{CLS}_i\} \oplus U_i$ is created and fed into \texttt{TE}, already trained by u-MLM, to create the embeddings $E_i = \{e^c_i, e^s_i, e^w_{i1}, .., e^w_{in}\}$.
The sequence $E^c = \{e^c_1, \ldots, e^c_n\}$ is fed into two transformer layers, \texttt{TL1} and \texttt{TL2}, that generate the new utterance embedding list $T^c = \{t^c_1, \ldots, t^c_n\}$.
Finally, $T^c$ is fed into a softmax layer that generates $o^\nu \in \mathbb{R}^{2}$ to predict whether or not $D'$ is in order.



\subsection{Fine-tuning for QA on Dialogue}
\label{ssec:finetuning}

Fine-tuning exploits multi-task learning between the utterance ID prediction (\textsec{sssec:finetuning-1}) and the token span prediction (\textsec{sssec:finetuning-2}), which allows the model to train both the utterance- and token-level attentions.
The transformer encoder (\texttt{TE}) trained by the utterance order prediction (UOP) is used for both tasks.
Given the question $Q = \{q_1, \ldots, q_n\}$  ($q_i$ is the $i$'th token in $Q$) and the dialogue $D = \{U_1, \ldots, U_m\}$, $Q$ and all $U_*$ are fed into \texttt{TE} that generates $E_q = \{e^c_q, e^q_1, .., e^q_n\}$ and $E_i = \{e^c_i, e^s_i, e^w_{i1}, .., e^w_{in}\}$ for $Q$ and every $U_i$, respectively.





\subsubsection{Utterance ID Prediction}
\label{sssec:finetuning-1}

The utterance embedding list $E^c = \{e^c_q, e^c_1, .., e^c_n\}$ is fed into \texttt{TL1} and \texttt{TL2} from UOP  that generate $T^c = \{t^c_q, t^c_1, .., t^c_n\}$.
$T^c$ is then fed into a softmax layer that generates $o^u \in \mathbb{R}^{m+1}$ to predict the ID of the utterance containing the answer span if exists; otherwise, the $0$'th label is predicted, implying that the answer span for $Q$ does not exist in $D$.



\subsubsection{Token Span Prediction}
\label{sssec:finetuning-2}

For every $E_i$, the pair $(E'_q, E'_i)$ is fed into the multi-head attention layer, \texttt{MHA}, where $E'_q = E_q \setminus \{e^c_q\}$ and $E'_i = E_i \setminus \{e^c_i\}$.
\texttt{MHA} \cite{vaswani_2017} then generates the attended embedding sequences, $T^a_1, \ldots, T^a_m$, where $T^a_i = \{t^s_i, t^w_{i1}, .., t^w_{in}\} $.
Finally, each $T^a_i$ is fed into two softmax layers, \texttt{SL} and \texttt{SR}, that generate $o^\ell_i \in \mathbb{R}^{n+1}$ and $o^r_i \in \mathbb{R}^{n+1}$ to predict\LN the leftmost and the rightmost tokens in $U_i$ respectively, that yield the answer span for $Q$.
It is possible that the answer spans are predicted in multiple utterances, in which case, the span from the utterance that has the highest score for the utterance ID prediction is selected, which is more efficient than the typical dynamic programming approach.


\section{Experiments}
\label{sec:experiments}

\subsection{Corpus}
\label{ssec:corpus}

Despite of all great work in QA, only two datasets are publicly available for machine comprehension that take dialogues as evidence documents.
One is \textsc{Dream} comprising dialogues for language exams with multiple-choice questions \cite{Sun_2019}.\LN
The other is \textsc{FriendsQA} containing transcripts from the TV show \textit{Friends} with\ annotation for span-based question answering \cite{yang_2019}.\LN
Since \textsc{Dream} is for a reading comprehension task that does not need to find the answer contents from the evidence documents, it is not suitable for our approach; thus, \textsc{FriendsQA} is chosen.

Each scene is treated as an independent dialogue in \textsc{FriendsQA}.
\citet{yang_2019} randomly split the corpus to generate training, development, and evaluation sets such that scenes from the same episode can be distributed across those three sets, causing inflated accuracy scores.
Thus, we re-split them by episodes to prevent such inflation.
For fine-tuning (\textsec{ssec:finetuning}), episodes from the first four seasons are used as described in Table~\ref{tbl:dataset-stats}.
For pre-training (\textsec{ssec:pretraining}), all transcripts from Seasons 5-10 are used as an additional training set.

\begin{table}[htbp!]
\centering\small
\begin{tabular}{c||r|r|r||l}
\bf Set & \multicolumn{1}{c|}{\bf D} & \multicolumn{1}{c|}{\bf Q} & \multicolumn{1}{c||}{\bf A} & \multicolumn{1}{c}{\bf E} \\
\hline\hline
Training    & 973 & 9,791 & 16,352 & $\:\:$1 - 20 \\
Development & 113 & 1,189 &  2,065 & 21 - 22 \\
Evaluation  & 136 & 1,172 &  1,920 & 23 - *  \\
\end{tabular}
\caption{New data split for \texttt{FriendsQA}. D/Q/A: \# of dialogues/questions/answers, E: episode IDs.}
\label{tbl:dataset-stats}
\vspace{-2ex}
\end{table}

\subsection{Models}
\label{ssec:models}

The weights from the \texttt{BERT}\textsubscript{base} and \texttt{RoBERTa}\textsubscript{base} models \cite{devlin_2019,liu_2019} are transferred to all models in our experiments.
Four baseline models, \texttt{BERT}, \texttt{BERT}\textsubscript{pre}, \texttt{RoBERTa}, and \texttt{RoBERTa}\textsubscript{pre}, are built, where all models are fine-tuned on the datasets in Table~\ref{tbl:dataset-stats} and the \texttt{*}\textsubscript{pre} models are pre-trained on the same datasets with the additional training set from Seasons 5-10 (\textsec{ssec:corpus}).
The baseline models are compared to \texttt{BERT}\textsubscript{our} and \texttt{RoBERTA}\textsubscript{our} that are trained by our approach.\footnote{Detailed experimental setup are provided in Appendices.}

\subsection{Results}

Table~\ref{tab:result} shows results achieved by all the models.
Following \citet{yang_2019}, exact matching (EM), span matching (SM), and utterance matching (UM) are used as the evaluation metrics.
Each model is developed three times and their average score as well as the standard deviation are reported. 
The performance of \texttt{RoBERTa*} is generally higher than \texttt{BERT*} although \texttt{RoBERTa}\textsubscript{base} is pre-trained with larger datasets including \textsc{CC-News} \cite{nagel_2016}, \textsc{OpenWebText} \cite{gokaslan_2019}, and \textsc{Stories} \cite{trinh_2018} than \texttt{BERT}\textsubscript{base} such that results from those two types of transformers cannot be directly compared.


\begin{table}[htp!]
\centering\resizebox{\columnwidth}{!}{
\begin{tabular}{l||c|c|c}
\multicolumn{1}{c|}{\bf Model} & \textbf{EM} & \textbf{SM} & \textbf{UM} \\
\hline \hline
\texttt{BERT}                    &         43.3($\pm$0.8)  &         59.3($\pm$0.6)  &         70.2($\pm$0.4) \\ 
\texttt{BERT}\textsubscript{pre} &         45.6($\pm$0.9)  &         61.2($\pm$0.7)  &         71.3($\pm$0.6) \\ 
\texttt{BERT}\textsubscript{our} & \textbf{46.8}($\pm$1.3) & \textbf{63.1}($\pm$1.1) & \textbf{73.3}($\pm$0.7) \\
\hline\hline
\texttt{RoBERTa}                    &         52.6($\pm$0.7)  &         68.2($\pm$0.3)  &         80.9($\pm$0.8) \\ 
\texttt{RoBERTa}\textsubscript{pre} &         52.6($\pm$0.7)  &         68.6($\pm$0.6)  &         81.7($\pm$0.7) \\ 
\texttt{RoBERTa}\textsubscript{our} & \textbf{53.5}($\pm$0.7) & \textbf{69.6}($\pm$0.8) & \textbf{82.7}($\pm$0.5) \\ 
\end{tabular}}
\caption{Accuracies ($\pm$ standard deviations) achieved by the \texttt{BERT} and \texttt{RoBERTa} models.}
\label{tab:result}
\end{table}

\noindent The \texttt{*}\textsubscript{pre} models show marginal improvement over their base models, implying that pre-training the language models on \textsc{FriendsQA} with the original transformers does not make much impact on this QA task. 
The models using our approach perform noticeably better than the baseline models, showing 3.8\% and 1.4\% improvements on SM from \texttt{BERT} and \texttt{RoBERTa}, respectively.

\begin{table}[htp!]
\centering\resizebox{\columnwidth}{!}{
\begin{tabular}{c|c||c|c|c}
\textbf{Type} &\textbf{Dist.} & \textbf{EM} & \textbf{SM} & \textbf{UM} \\
\hline \hline
\tt Where & 18.16 & 66.1($\pm$0.5) & 79.9($\pm$0.7) & 89.8($\pm$0.7) \\
\tt When  & 13.57 & 63.3($\pm$1.3) & 76.4($\pm$0.6) & 88.9($\pm$1.2) \\
\tt What  & 18.48 & 56.4($\pm$1.7) & 74.0($\pm$0.5) & 87.7($\pm$2.1) \\
\tt Who   & 18.82 & 55.9($\pm$0.8) & 66.0($\pm$1.7) & 79.9($\pm$1.1) \\
\tt How   & 15.32 & 43.2($\pm$2.3) & 63.2($\pm$2.5) & 79.4($\pm$0.7) \\
\tt Why   & 15.65 & 33.3($\pm$2.0) & 57.3($\pm$0.8) & 69.8($\pm$1.8) \\
\end{tabular}}
\caption{Results from the \texttt{RoBERTa}\textsubscript{our} model by different question types.}
\label{tab:question-type}
\end{table}

\noindent Table~\ref{tab:question-type} shows the results achieved by \texttt{RoBERTa}\textsubscript{our} w.r.t.\ question types.
UM drops significantly for \texttt{Why} that often spans out to longer sequences and also requires deeper inferences to answer correctly than the others.
Compared to the baseline models, our models show more well-around performance regardless the question types.\footnote{Question type results for all models are in Appendices.}

\subsection{Ablation Studies}

Table~\ref{tab:ablation} shows the results from ablation studies to analyze the impacts of the individual approaches.
\texttt{BERT}\textsubscript{pre} and \texttt{RoBERTa}\textsubscript{pre} are the same as in Table~\ref{tab:result}, that are the transformer models pre-trained by the token-level masked LM (\textsec{sssec:pretraining-1}) and fine-tuned by the token span prediction (\textsec{sssec:finetuning-2}).
\texttt{BERT}\textsubscript{uid} and \texttt{RoBERTa}\textsubscript{uid} are the models that are pre-trained by the token-level masked LM and jointly fine-tuned by the token span prediction as well as the utterance ID prediction (UID: \textsec{sssec:finetuning-1}).
Given these two types of transformer models, the utterance-level masked LM (ULM: \textsec{sssec:pretraining-2}) and the utterance order prediction (UOP: \textsec{sssec:pretraining-3}) are separately evaluated.

\begin{table}[htp!]
\centering\resizebox{\columnwidth}{!}{
\begin{tabular}{l||c|c|c}
\multicolumn{1}{c||}{\bf Model} & \bf EM & \bf SM & \bf UM  \\
\hline \hline
\texttt{BERT}\textsubscript{pre}  & 45.6($\pm$0.9) & 61.2($\pm$0.7) & 71.3($\pm$0.6) \\
\textsc{$\oplus$ulm}              & 45.7($\pm$0.9) & 61.8($\pm$0.9) & 71.8($\pm$0.5) \\
\textsc{$\oplus$ulm$\oplus$uop}   & 45.6($\pm$0.9) & 61.7($\pm$0.7) & 71.7($\pm$0.6) \\
\hline
\texttt{BERT}\textsubscript{uid}  & 45.7($\pm$0.8) & 61.1($\pm$0.8) & 71.5($\pm$0.5) \\
\textsc{$\oplus$ulm}              & 46.2($\pm$1.1) & 62.4($\pm$1.2) & 72.5($\pm$0.8) \\
\textsc{$\oplus$ulm$\oplus$uop}   & \textbf{46.8}($\pm$1.3) & \textbf{63.1}($\pm$1.1) & \textbf{73.3}($\pm$0.7) \\
\hline \hline
\texttt{RoBERTa}\textsubscript{pre} & 52.6($\pm$0.7) & 68.6($\pm$0.6) & 81.7($\pm$0.7) \\
\textsc{$\oplus$ulm}                & 52.9($\pm$0.8) & 68.7($\pm$1.1) & 81.7($\pm$0.6) \\
\textsc{$\oplus$ulm$\oplus$uop}     & 52.5($\pm$0.8) & 68.8($\pm$0.5) & 81.9($\pm$0.7) \\
\hline
\texttt{RoBERTa}\textsubscript{uid} & 52.8($\pm$0.9) & 68.7($\pm$0.8) & 81.9($\pm$0.5) \\
\textsc{$\oplus$ulm}                & 53.2($\pm$0.6) & 69.2($\pm$0.7) & 82.4($\pm$0.5) \\
\textsc{$\oplus$ulm$\oplus$uop}     & \textbf{53.5}($\pm$0.7) & \textbf{69.6}($\pm$0.8) & \textbf{82.7}($\pm$0.5) \\
\end{tabular}}
\caption{Results for the ablation studies. Note that the \texttt{*}\textsubscript{uid}$\oplus$\textsc{ulm}$\oplus$\textsc{uop} models are equivalent to the \texttt{*}\textsubscript{our} models in Table~\ref{tab:result}, respectively.}
\label{tab:ablation}
\end{table}

\noindent These two dialogue-specific LM approaches, ULM and UOP, give very marginal improvement over the baseline models, that is rather surprising.
However, they show good improvement when combined with UID, implying that pre-training language models may not be enough to enhance the performance by itself but can be effective when it is coupled with an appropriate fine-tuning approach.
Since both ULM and UOP are designed to improve the quality of utterance embeddings, it is expected to improve the accuracy for UID as well. 
The improvement on UM is indeed encouraging, giving 2\% and 1\% boosts to \texttt{BERT}\textsubscript{pre} and \texttt{RoBERTa}\textsubscript{pre}, respectively and consequently improving the other two metrics.


\subsection{Error Analysis}

As shown in Table~\ref{tab:question-type}, the major errors are from the three types of questions, \texttt{who}, \texttt{how}, and \texttt{why}; thus, we select 100 dialogues associated with those question types that our best model, \texttt{RoBERTa}\textsubscript{our}, incorrectly predicts the answer spans for.
Specific examples are provided in Tables \ref{tab:error_why}, \ref{tab:error_who} and \ref{tab:error_how} (\textsec{sup:error-examples}).\LN
Following \citet{yang_2019a}, errors are grouped into 6 categories, entity resolution, paraphrase and partial match, cross-utterance reasoning, question bias, noise in annotation, and miscellaneous.

\noindent Table~\ref{tab:error_types} shows the errors types and their ratios with respect to the question types.
Two main error types are entity resolution and cross-utterance reasoning.
The entity resolution error happens when many of the same entities are mentioned in multiple utterances. 
This error also occurs when the QA system is asked about a specific person, but predicts wrong people where there are so many people appearing in multiple utterances. 
The cross-utterance reasoning error often happens with the \texttt{why} and \texttt{how} questions where the model relies on pattern matching mostly and predicts the next utterance span of the matched pattern.

\begin{table}[htp!]
\centering\resizebox{\columnwidth}{!}{\begin{tabular}{c||c|c|c}
\bf Error Types & \bf\texttt{Who} & \bf\texttt{How} & \bf\texttt{Why}  \\
\hline \hline
Entity Resolution            & \textbf{34\%} & 23\% & 20\% \\
Paraphrase and Partial Match & 14\% & 14\% & 13\% \\
Cross-Utterance Reasoning    & 25\% & \textbf{28\%} & \textbf{27\%} \\
Question Bias                & 11\% & 13\% & 17\% \\
Noise in Annotation          & 4\%  & 7\%  & 9\%  \\
Miscellaneous                & 12\% & 15\% & 14\%
\end{tabular}}
\caption{Error types and their ratio with respect to the three most challenging question types.}
\label{tab:error_types}
\end{table}
\section{Conclusion}
This paper introduces a novel transformer approach that effectively interprets hierarchical contexts in multiparty dialogue by learning utterance embeddings.
Two language modeling approaches are proposed, utterance-level masked LM and utterance order prediction.
Coupled with the joint inference between token span prediction and utterance ID prediction, these two language models significantly outperform two of the state-of-the-art transformer approaches, \texttt{BERT} and \texttt{RoBERTa}, on a span-based QA task called \textit{FriendsQA}   .
We will evaluate our approach on other machine comprehension tasks using dialogues as evidence documents to further verify the generalizability of this work.



\label{sec:conclusion}
\section*{Acknowledgments}

We gratefully acknowledge the support of the AWS Machine Learning Research Awards (MLRA).
Any contents in this material are those of the authors and do not necessarily reflect the views of them.

\bibliography{acl2020}

\begin{thebibliography}{21}
\expandafter\ifx\csname natexlab\endcsname\relax\def\natexlab#1{#1}\fi

\bibitem[{Choi et~al.(2018)Choi, He, Iyyer, Yatskar, Yih, Choi, Liang, and
  Zettlemoyer}]{Choi_2018}
Eunsol Choi, He~He, Mohit Iyyer, Mark Yatskar, Wen-tau Yih, Yejin Choi, Percy
  Liang, and Luke Zettlemoyer. 2018.
\newblock \href {https://doi.org/10.18653/v1/d18-1241} {Quac: Question
  answering in context}.
\newblock \emph{Proceedings of the 2018 Conference on Empirical Methods in
  Natural Language Processing}.

\bibitem[{CONNEAU and Lample(2019)}]{lample_2019}
Alexis CONNEAU and Guillaume Lample. 2019.
\newblock \href
  {http://papers.nips.cc/paper/8928-cross-lingual-language-model-pretraining.pdf}
  {Cross-lingual language model pretraining}.
\newblock In H.~Wallach, H.~Larochelle, A.~Beygelzimer, F.~d\textquotesingle
  Alch\'{e}-Buc, E.~Fox, and R.~Garnett, editors, \emph{Advances in Neural
  Information Processing Systems 32}, pages 7057--7067. Curran Associates, Inc.

\bibitem[{Devlin et~al.(2019)Devlin, Chang, Lee, and Toutanova}]{devlin_2019}
Jacob Devlin, Ming-Wei Chang, Kenton Lee, and Kristina Toutanova. 2019.
\newblock \href {https://www.aclweb.org/anthology/N19-1423} {{BERT:
  Pre-training of Deep Bidirectional Transformers for Language Understanding}}.
\newblock In \emph{Proceedings of the 2019 Conference of the North {A}merican
  Chapter of the Association for Computational Linguistics: Human Language
  Technologies}, NAACL'19, pages 4171--4186.

\bibitem[{Gokaslan and Cohen(2019)}]{gokaslan_2019}
Aaron Gokaslan and Vanya Cohen. 2019.
\newblock \href {https://skylion007.github.io/OpenWebTextCorpus/}
  {\emph{{OpenWebText Corpus}}}.

\bibitem[{Iyyer et~al.(2017)Iyyer, Yih, and Chang}]{iyyer_2017}
Mohit Iyyer, Wen-tau Yih, and Ming-Wei Chang. 2017.
\newblock \href {https://doi.org/10.18653/v1/P17-1167} {Search-based neural
  structured learning for sequential question answering}.
\newblock In \emph{Proceedings of the 55th Annual Meeting of the Association
  for Computational Linguistics (Volume 1: Long Papers)}, pages 1821--1831,
  Vancouver, Canada. Association for Computational Linguistics.

\bibitem[{Joshi et~al.(2017)Joshi, Choi, Weld, and Zettlemoyer}]{Joshi_2017}
Mandar Joshi, Eunsol Choi, Daniel Weld, and Luke Zettlemoyer. 2017.
\newblock \href {https://doi.org/10.18653/v1/p17-1147} {Triviaqa: A large scale
  distantly supervised challenge dataset for reading comprehension}.
\newblock \emph{Proceedings of the 55th Annual Meeting of the Association for
  Computational Linguistics (Volume 1: Long Papers)}.

\bibitem[{Kočiský et~al.(2018)Kočiský, Schwarz, Blunsom, Dyer, Hermann,
  Melis, and Grefenstette}]{Ko_isk__2018}
Tomáš Kočiský, Jonathan Schwarz, Phil Blunsom, Chris Dyer, Karl~Moritz
  Hermann, Gábor Melis, and Edward Grefenstette. 2018.
\newblock \href {https://doi.org/10.1162/tacl_a_00023} {The narrativeqa reading
  comprehension challenge}.
\newblock \emph{Transactions of the Association for Computational Linguistics},
  6:317–328.

\bibitem[{Lan et~al.(2019)Lan, Chen, Goodman, Gimpel, Sharma, and
  Soricut}]{lan_2019}
Zhenzhong Lan, Mingda Chen, Sebastian Goodman, Kevin Gimpel, Piyush Sharma, and
  Radu Soricut. 2019.
\newblock \href {http://arxiv.org/abs/1909.11942} {Albert: A lite bert for
  self-supervised learning of language representations}.

\bibitem[{Liu et~al.(2019)Liu, Ott, Goyal, Du, Joshi, Chen, Levy, Lewis,
  Zettlemoyer, and Stoyanov}]{liu_2019}
Yinhan Liu, Myle Ott, Naman Goyal, Jingfei Du, Mandar Joshi, Danqi Chen, Omer
  Levy, Mike Lewis, Luke Zettlemoyer, and Veselin Stoyanov. 2019.
\newblock \href {http://arxiv.org/abs/1907.11692} {{RoBERTa: A Robustly
  Optimized BERT Pretraining Approach}}.
\newblock \emph{arXiv}, 1907.11692.

\bibitem[{Nagel(2016)}]{nagel_2016}
Sebastian Nagel. 2016.
\newblock \href {https://commoncrawl.org/2016/10/news-dataset-available/}
  {\emph{{News Dataset Available}}}.

\bibitem[{Nguyen et~al.(2016)Nguyen, Rosenberg, Song, Gao, Tiwary, Majumder,
  and Deng}]{bajaj_2016}
Tri Nguyen, Mir Rosenberg, Xia Song, Jianfeng Gao, Saurabh Tiwary, Rangan
  Majumder, and Li~Deng. 2016.
\newblock \href {http://ceur-ws.org/Vol-1773/CoCoNIPS\_2016\_paper9.pdf} {{MS}
  {MARCO:} {A} human generated machine reading comprehension dataset}.
\newblock In \emph{Proceedings of the Workshop on Cognitive Computation:
  Integrating neural and symbolic approaches 2016 co-located with the 30th
  Annual Conference on Neural Information Processing Systems {(NIPS} 2016),
  Barcelona, Spain, December 9, 2016}.

\bibitem[{Rajpurkar et~al.(2018)Rajpurkar, Jia, and Liang}]{Rajpurkar_2018}
Pranav Rajpurkar, Robin Jia, and Percy Liang. 2018.
\newblock \href {https://doi.org/10.18653/v1/p18-2124} {Know what you don’t
  know: Unanswerable questions for squad}.
\newblock \emph{Proceedings of the 56th Annual Meeting of the Association for
  Computational Linguistics (Volume 2: Short Papers)}.

\bibitem[{Rajpurkar et~al.(2016)Rajpurkar, Zhang, Lopyrev, and
  Liang}]{Rajpurkar_2016}
Pranav Rajpurkar, Jian Zhang, Konstantin Lopyrev, and Percy Liang. 2016.
\newblock \href {https://doi.org/10.18653/v1/d16-1264} {Squad: 100,000+
  questions for machine comprehension of text}.
\newblock \emph{Proceedings of the 2016 Conference on Empirical Methods in
  Natural Language Processing}.

\bibitem[{Reddy et~al.(2019)Reddy, Chen, and Manning}]{Reddy_2019}
Siva Reddy, Danqi Chen, and Christopher~D. Manning. 2019.
\newblock \href {https://doi.org/10.1162/tacl_a_00266} {Coqa: A conversational
  question answering challenge}.
\newblock \emph{Transactions of the Association for Computational Linguistics},
  7:249–266.

\bibitem[{Sun et~al.(2019)Sun, Yu, Chen, Yu, Choi, and Cardie}]{Sun_2019}
Kai Sun, Dian Yu, Jianshu Chen, Dong Yu, Yejin Choi, and Claire Cardie. 2019.
\newblock \href {https://www.aclweb.org/anthology/Q19-1014} {{DREAM: A
  Challenge Data Set and Models for Dialogue-Based Reading Comprehension}}.
\newblock \emph{Transactions of the Association for Computational Linguistics},
  7:217--231.

\bibitem[{Talmor and Berant(2018)}]{Talmor_2018}
Alon Talmor and Jonathan Berant. 2018.
\newblock \href {https://doi.org/10.18653/v1/n18-1059} {The web as a
  knowledge-base for answering complex questions}.
\newblock \emph{Proceedings of the 2018 Conference of the North American
  Chapter of the Association for Computational Linguistics: Human Language
  Technologies, Volume 1 (Long Papers)}.

\bibitem[{Trinh and Le(2018)}]{trinh_2018}
Trieu~H. Trinh and Quoc~V. Le. 2018.
\newblock \href {http://arxiv.org/abs/1806.02847} {{A Simple Method for
  Commonsense Reasoning}}.
\newblock \emph{arXiv}, 1806.02847.

\bibitem[{Trischler et~al.(2017)Trischler, Wang, Yuan, Harris, Sordoni,
  Bachman, and Suleman}]{Trischler_2017}
Adam Trischler, Tong Wang, Xingdi Yuan, Justin Harris, Alessandro Sordoni,
  Philip Bachman, and Kaheer Suleman. 2017.
\newblock \href {https://doi.org/10.18653/v1/w17-2623} {Newsqa: A machine
  comprehension dataset}.
\newblock \emph{Proceedings of the 2nd Workshop on Representation Learning for
  NLP}.

\bibitem[{Vaswani et~al.(2017)Vaswani, Shazeer, Parmar, Uszkoreit, Jones,
  Gomez, Kaiser, and Polosukhin}]{vaswani_2017}
Ashish Vaswani, Noam Shazeer, Niki Parmar, Jakob Uszkoreit, Llion Jones,
  Aidan~N. Gomez, Lukasz Kaiser, and Illia Polosukhin. 2017.
\newblock \href {http://dl.acm.org/citation.cfm?id=3295222.3295349} {Attention
  is all you need}.
\newblock In \emph{Proceedings of the 31st International Conference on Neural
  Information Processing Systems}, NIPS'17, pages 6000--6010, USA. Curran
  Associates Inc.

\bibitem[{Yang and Choi(2019)}]{yang_2019}
Zhengzhe Yang and Jinho~D. Choi. 2019.
\newblock \href {https://www.aclweb.org/anthology/W19-5923} {{F}riends{QA}:
  Open-domain question answering on {TV} show transcripts}.
\newblock In \emph{Proceedings of the 20th Annual SIGdial Meeting on Discourse
  and Dialogue}, pages 188--197, Stockholm, Sweden. Association for
  Computational Linguistics.

\bibitem[{Yang et~al.(2019)Yang, Dai, Yang, Carbonell, Salakhutdinov, and
  Le}]{yang_2019a}
Zhilin Yang, Zihang Dai, Yiming Yang, Jaime Carbonell, Russ~R Salakhutdinov,
  and Quoc~V Le. 2019.
\newblock \href
  {http://papers.nips.cc/paper/8812-xlnet-generalized-autoregressive-pretraining-for-language-understanding.pdf}
  {Xlnet: Generalized autoregressive pretraining for language understanding}.
\newblock In H.~Wallach, H.~Larochelle, A.~Beygelzimer, F.~d\textquotesingle
  Alch\'{e}-Buc, E.~Fox, and R.~Garnett, editors, \emph{Advances in Neural
  Information Processing Systems 32}, pages 5754--5764. Curran Associates, Inc.

\end{thebibliography}
\bibliographystyle{acl_natbib}

\cleardoublepage\appendix
\section{Appendices}
\label{sec:supplemental-materials}

\subsection{Experimental Setup}
\label{sup:experimental-setup}

The \texttt{BERT}$_{{\rm base}}$ model and the \texttt{RoBERTa}$_{{\rm BASE}}$ model use the same configuration. The two models both have 12 hidden transformer layers and 12 attention heads. The hidden size of the model is 768 and the intermediate size in the transformer layers is 3,072. The activation function in the transformer layers is \texttt{gelu}.

\paragraph{Pre-training} 

The batch size of 32 sequences is used for pre-training.
\texttt{Adam} with the learning rate of $5\cdot 10^{-5}$, $\beta_1 = 0.9$, $\beta_2 = 0.999$, the \texttt{L2} weight decay of $0.01$, the learning rate warm up over the first 10\% steps, and the linear decay of the learning rate are used.
A dropout probability of $0.1$ is applied to all layers.
The cross-entropy is used for the training loss of each task.
For the masked language modeling tasks, the model is trained until the perplexity stops decreasing on the development set. For the other pre-training tasks, the model is trained until both the loss and the accuracy stop decreasing on the development set.

\paragraph{Fine-tuning}

For fine-tuning, the batch size and the optimization approach are the same as the pre-training.
The dropout probability is always kept at $0.1$. The training loss is the sum of the cross-entropy of two fine-tuning tasks as in \textsec{ssec:finetuning}.

\subsection{Question Types Analysis}
\label{sup:question-type-analysis}

Tables in this section show the results with respect to the question types using all models (Section~\ref{ssec:models}) in the order of performance. 

\begin{table}[htp!]
\centering\resizebox{\columnwidth}{!}{
\begin{tabular}{c|c||c|c|c}
\textbf{Type} &\textbf{Dist.} & \textbf{EM} & \textbf{SM} & \textbf{UM} \\
\hline \hline
\tt Where & 18.16 & 68.3($\pm$1.3) & 78.8($\pm$1.2) & 89.2($\pm$1.5) \\
\tt When  & 13.57 & 63.8($\pm$1.6) & 75.2($\pm$0.9) & 86.0($\pm$1.6) \\
\tt What  & 18.48 & 54.1($\pm$0.8) & 72.5($\pm$1.5) & 84.0($\pm$0.9) \\
\tt Who   & 18.82 & 56.0($\pm$1.3) & 66.1($\pm$1.3) & 79.4($\pm$1.2) \\
\tt How   & 15.32 & 38.1($\pm$0.7) & 59.2($\pm$1.6) & 77.5($\pm$0.7) \\
\tt Why   & 15.65 & 32.0($\pm$1.1) & 56.0($\pm$1.7) & 68.5($\pm$0.8) \\
\end{tabular}}
\caption{Results from \texttt{RoBERTa} by question types.}
\label{tab:types_result_for_RoBERTa}
\end{table}

\begin{table}[htp!]
\centering\resizebox{\columnwidth}{!}{
\begin{tabular}{c|c||c|c|c}
\textbf{Type} &\textbf{Dist.} & \textbf{EM} & \textbf{SM} & \textbf{UM} \\
\hline \hline
\tt Where & 18.16 & 67.1($\pm$1.2) & 78.9($\pm$0.6) & 89.0($\pm$1.1) \\
\tt When  & 13.57 & 62.3($\pm$0.7) & 76.3($\pm$1.3) & 88.7($\pm$0.9) \\
\tt What  & 18.48 & 55.1($\pm$0.8) & 73.1($\pm$0.8) & 86.7($\pm$0.8) \\
\tt Who   & 18.82 & 56.2($\pm$1.4) & 64.0($\pm$1.7) & 77.1($\pm$1.3) \\
\tt How   & 15.32 & 41.2($\pm$1.1) & 61.2($\pm$1.5) & 79.8($\pm$0.7) \\
\tt Why   & 15.65 & 32.4($\pm$0.7) & 57.4($\pm$0.8) & 69.1($\pm$1.4) \\
\end{tabular}}
\caption{Results from \texttt{RoBERTa}\textsubscript{pre} by question types.}
\label{tab:types_result_for_RoBERTa_pre}
\end{table}

\begin{table}[htp!]
\centering\resizebox{\columnwidth}{!}{
\begin{tabular}{c|c||c|c|c}
\textbf{Type} &\textbf{Dist.} & \textbf{EM} & \textbf{SM} & \textbf{UM} \\
\hline \hline
\tt Where & 18.16 & 66.1($\pm$0.5) & 79.9($\pm$0.7) & 89.8($\pm$0.7) \\
\tt When  & 13.57 & 63.3($\pm$1.3) & 76.4($\pm$0.6) & 88.9($\pm$1.2) \\
\tt What  & 18.48 & 56.4($\pm$1.7) & 74.0($\pm$0.5) & 87.7($\pm$2.1) \\
\tt Who   & 18.82 & 55.9($\pm$0.8) & 66.0($\pm$1.7) & 79.9($\pm$1.1) \\
\tt How   & 15.32 & 43.2($\pm$2.3) & 63.2($\pm$2.5) & 79.4($\pm$0.7) \\
\tt Why   & 15.65 & 33.3($\pm$2.0) & 57.3($\pm$0.8) & 69.8($\pm$1.8) \\
\end{tabular}}
\caption{Results from \texttt{RoBERTa}\textsubscript{our} by question types.}
\label{tab:types_result_for_RoBERTa_Our}
\end{table}

\begin{table}[htp!]
\centering\resizebox{\columnwidth}{!}{
\begin{tabular}{c|c||c|c|c}
\textbf{Type} &\textbf{Dist.} & \textbf{EM} & \textbf{SM} & \textbf{UM} \\
\hline \hline
\tt Where & 18.16 & 57.3($\pm$0.5) & 70.2($\pm$1.3) & 79.4($\pm$0.9) \\
\tt When  & 13.57 & 56.1($\pm$1.1) & 69.7($\pm$1.6) & 78.6($\pm$1.7) \\
\tt What  & 18.48 & 45.0($\pm$1.4) & 64.4($\pm$0.7) & 77.0($\pm$1.0) \\
\tt Who   & 18.82 & 46.9($\pm$1.1) & 56.2($\pm$1.4) & 67.6($\pm$1.4) \\
\tt How   & 15.32 & 29.3($\pm$0.8) & 48.4($\pm$1.2) & 60.9($\pm$0.7) \\
\tt Why   & 15.65 & 23.4($\pm$1.6) & 46.1($\pm$0.9) & 56.4($\pm$1.3) \\
\end{tabular}}
\caption{Results from \texttt{BERT} by question types.}
\label{tab:types_result_for_BERT}
\end{table}

\begin{table}[htp!]
\centering\resizebox{\columnwidth}{!}{
\begin{tabular}{c|c||c|c|c}
\textbf{Type} &\textbf{Dist.} & \textbf{EM} & \textbf{SM} & \textbf{UM} \\
\hline \hline
\tt Where & 18.16 & 62.8($\pm$1.8) & 72.3($\pm$0.8) & 82.1($\pm$0.7) \\
\tt When  & 13.57 & 60.7($\pm$1.5) & 70.7($\pm$1.8) & 80.4($\pm$1.1) \\
\tt What  & 18.48 & 43.2($\pm$1.3) & 64.3($\pm$1.7) & 75.6($\pm$1.8) \\
\tt Who   & 18.82 & 47.8($\pm$1.1) & 56.9($\pm$1.9) & 69.7($\pm$0.7) \\
\tt How   & 15.32 & 33.2($\pm$1.3) & 48.3($\pm$0.6) & 59.8($\pm$1.1) \\
\tt Why   & 15.65 & 22.9($\pm$1.6) & 46.6($\pm$0.7) & 54.9($\pm$0.9) \\
\end{tabular}}
\caption{Results from \texttt{BERT}\textsubscript{pre} by question types.}
\label{tab:types_result_for_BERT_pre}
\vspace{-2ex}
\end{table}

\begin{table}[htp!]
\centering\resizebox{\columnwidth}{!}{
\begin{tabular}{c|c||c|c|c}
\textbf{Type} &\textbf{Dist.} & \textbf{EM} & \textbf{SM} & \textbf{UM} \\
\hline \hline
\tt Where & 18.16 & 63.3($\pm$1.2) & 72.9($\pm$1.7) & 77.0($\pm$1.2) \\
\tt When  & 13.57 & 48.4($\pm$1.9) & 66.5($\pm$0.8) & 79.5($\pm$1.5) \\
\tt What  & 18.48 & 52.1($\pm$0.7) & 69.2($\pm$1.1) & 81.3($\pm$0.7) \\
\tt Who   & 18.82 & 51.3($\pm$1.1) & 61.9($\pm$0.9) & 67.5($\pm$0.9) \\
\tt How   & 15.32 & 30.9($\pm$0.9) & 52.1($\pm$0.7) & 65.4($\pm$1.1) \\
\tt Why   & 15.65 & 29.2($\pm$1.6) & 53.2($\pm$1.3) & 65.7($\pm$0.8) \\
\end{tabular}}
\caption{Results from \texttt{BERT}\textsubscript{our} by question types.}
\label{tab:types_result_for_BERT_our}
\end{table}

\subsection{Error Examples}
\label{sup:error-examples}

Each table in this section gives an error example from the excerpt.
The gold answers are indicated by the \uline{solid} underlines whereas the predicted answers are indicated by the \uwave{wavy} underlines. 

\begin{table}[htp!]
\centering\resizebox{\columnwidth}{!}{\begin{tabular}{c|l}
\textbf{Q} & \multicolumn{1}{c}{\textbf{Why is Joey planning a big party?}} \\
\hline\hline
J & Oh, \uwave{we're having a big party tomorrow night.} Later! \\
R & Whoa! Hey-hey, you planning on inviting us? \\
J & Nooo, later. \\
P & Hey!! Get your ass back here, Tribbiani!! \\
R & Hormones! \\
M & What Phoebe meant to say was umm, how come \\
  & you're having a party and we're not invited? \\
J & Oh, \uline{it's Ross' bachelor party.} \\
M & Sooo? \\
\end{tabular}}
\caption{An error example for the \texttt{why} question (Q).\\J: Joey, R: Rachel, P: Pheobe, M: Monica.}
\label{tab:error_why}
\end{table}

\begin{table}[htp!]
\centering\resizebox{\columnwidth}{!}{\begin{tabular}{c|l}
\textbf{Q} & \multicolumn{1}{c}{\textbf{Who opened the vent?}} \\
\hline\hline
\uline{R} & Ok, got the vent open. \\
P & Hi, I'm Ben. I'm hospital worker Ben. \\
  & It's Ben... to the rescue! \\
R & Ben, you ready? All right, gimme your foot. \\
  & Ok, on three, Ben. One, two, three. Ok, That's it, Ben. \\
- & (\textit{Ross and Susan lift Phoebe up into the vent}.) \\
S & What do you see? \\
P & Well, Susan, I see what appears to be a dark vent. \\
  & Wait. Yes, it is in fact a dark vent. \\
- & (\textit{\uwave{A janitor} opens the closet door from the outside.}) \\
\end{tabular}}
\caption{An error example for the \texttt{who} question (Q).\\P: Pheobe, R: Ross, S: Susan.}
\label{tab:error_who}
\end{table}

\begin{table}[htp!]
\centering\resizebox{\columnwidth}{!}{\begin{tabular}{c|l}
\multirow{2}{*}{\bf Q} & \multicolumn{1}{c}{\textbf{How does Joey try to convince the girl }} \\
                       & \multicolumn{1}{c}{\textbf{to hang out with him?}} \\
\hline\hline
J & Oh yeah-yeah. And I got the duck totally trained. \\
  & Watch this. Stare at the wall. Hardly move. Be white. \\
G & You are really good at that. \\
  & So uh, I had fun tonight, you throw one hell of a party. \\
J & Oh thanks. Thanks. It was great meetin' ya. And listen \\
  & if any of my friends gets married, or have a birthday, ... \\
G & Yeah, that would be great. So I guess umm, good night. \\
J & \uwave{Oh unless you uh, you wanna hang around.} \\
G & Yeah? \\
J & Yeah. \uline{I'll let you play with my duck.} \\
\end{tabular}}
\caption{An error example for the \texttt{how} question (Q).\\J: Joey, G: The Girl.}
\label{tab:error_how}
\end{table}

\end{document}